\documentclass[letterpaper, 10 pt, conference]{ieeeconf}  
\usepackage{xcolor}

\usepackage{cite}
\usepackage{graphicx}
\usepackage{float}
\usepackage{eurosym}
\usepackage{caption}
\usepackage{subcaption}
\usepackage{array}
\usepackage{seqsplit}
\usepackage{hyperref}
\newcolumntype{L}[1]{>{\raggedright\arraybackslash}p{#1}}
\usepackage{tikz}
\usepackage{booktabs}
\usepackage{tabularx}
\usepackage{subcaption}
\usepackage[inkscapelatex=false]{svg}

\IEEEoverridecommandlockouts                              

\title{\LARGE \bf AIfred: Augmented Learning through Functional Robotic Embodiment at the Desk}

\author{Gregorio Orlando, Milan Groshev, Eduardo Castell{\'{o}} Ferrer\\
\normalsize{CyPhy Life, Robotics \& AI Lab, School of Science \& Technology, IE University, Spain}}

\begin{document}

\maketitle
\thispagestyle{empty}
\pagestyle{empty}

\begin{abstract}
Desk-based learning and creative activities benefit from handwritten engagement. However, current generative AI tools deliver guidance through a separate screen, creating a gap between where users think and where assistance appears. To address this, in this work we design AIfred, a desk-based robotic arm with a projector mounted at the end-effector that places AI-generated guidance alongside handwritten work. AIfred combines workspace perception, context-aware content generation, and robot-mediated projection to support math-assignments, image-generation, and drawing tasks. In a user study (n = 36), we compared AIfred against ChatGPT (GPT-5.6 Luna) running on a laptop. Both tools performed comparably while assistance was available during math assignment (6.7 vs. 7.3/10, p = .41), but AIfred improved short-term learning transfer by 60\% once assistance was withdrawn (7.0 vs. 4.4/10, p = .003). In addition, independent art and design professors ranked drawings produced with AIfred better in 33 of 36 cases. Our findings indicate that spatially co-located AI assistance benefits tasks whose guidance shares a spatial frame with the work.
\end{abstract}

\section{INTRODUCTION}
Currently, many learning and creative activities such as solving problems or sketching new ideas unfold on the physical desk. Research shows that students who take handwritten notes outperform those using laptops on conceptual understanding. This is because handwriting encourages reformulation on paper and screen-based tools are more prone to passive transcription~\cite{mueller2014pen, OECD2026PISA}. Despite this, current AI tools are used exclusively through screens: users type prompts and receive a textual or visual response. Then, the resulting information is transferred back to the physical workspace (e.g., drawing a picture, or writing an essay). This workflow creates a disconnect (a.k.a., context switching) where the user works on paper while guidance lives on a screen. 

Spatial Augmented Reality (SAR) (i.e., a technique that projects virtual images onto real objects) offers a way to close this gap by embedding digital information directly into physical workspaces~\cite{bimber2005spatial},~\cite{wellner1993digitaldesk},~\cite{ullmer1997metadesk}. 
More recently, robotic projection platforms like LuminAR~ \cite{linder2010luminar} and ELEGNT~\cite{hu2025elegnt} introduced a robotic desk companion that enables interactive projection on arbitrary surfaces. However, these systems primarily use robotic embodiment to reposition projected content or to communicate through expressive motion. They do not perceive the user's physical task, generate context-dependent learning support, and project that guidance back into the workspace where the task is being performed.

\begin{figure}[!t]
    \centering
    \includegraphics[width=0.8\columnwidth]{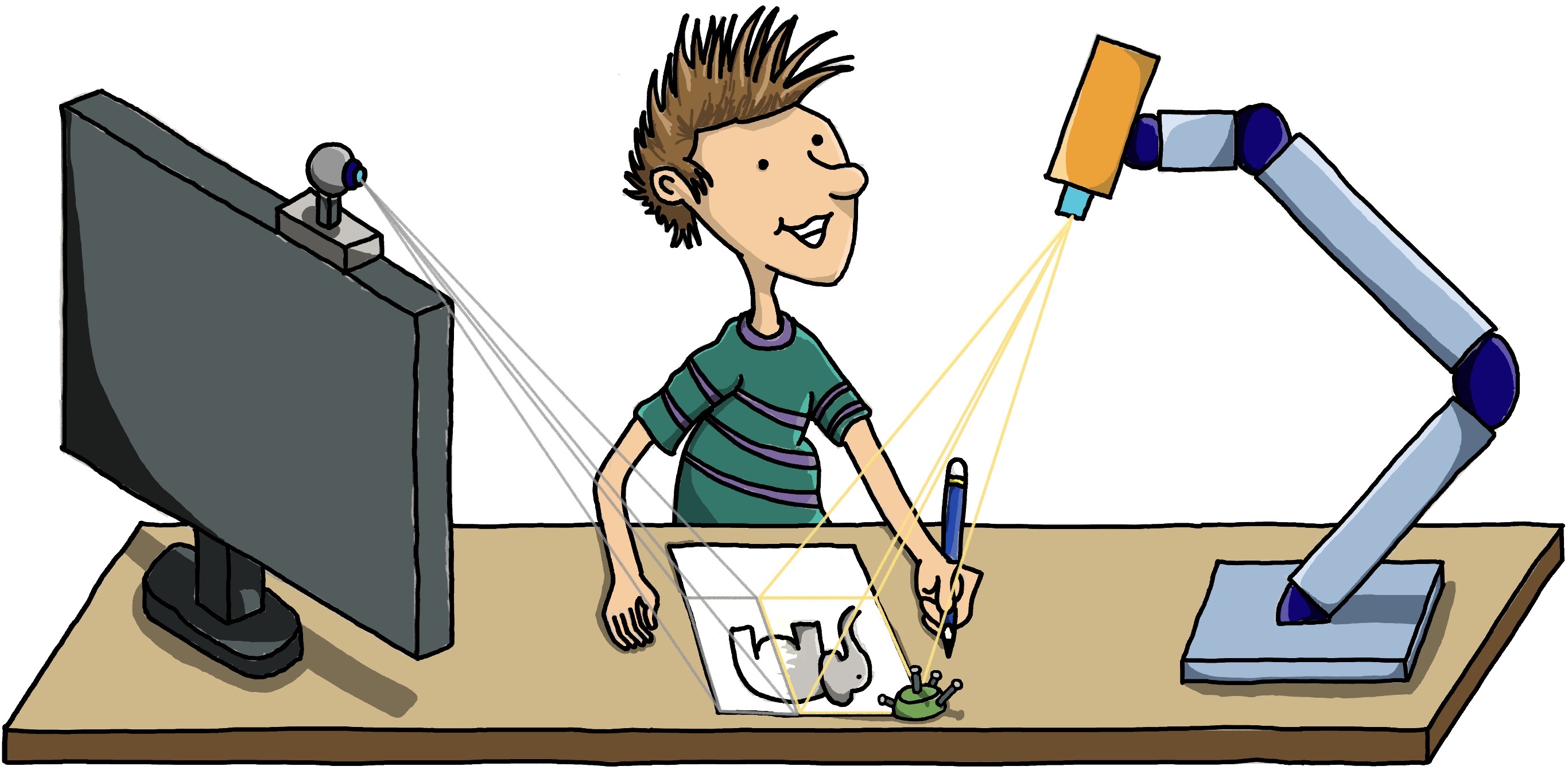}
    \caption{AIfred combines a robotic arm, a projector, and an overhead camera for workspace perception to deliver AI assistance directly within the user’s physical desk.}
    \label{fig:AIfredSketch}
    \vspace{-5mm}
\end{figure}

In this work, we present AIfred, a desk-based robotic arm with a mini projector mounted at its end-effector. An overhead camera captures the user's workspace, while a motion-capture system tracks an object that the user moves to indicate where projected content should appear (see Fig.~\ref{fig:AIfredSketch}). This setup allows the system to reposition AI-generated content anywhere on the desk, keeping digital guidance physically aligned with the user's handwritten work.

AIfred operates through a spatial assistance pipeline comprising three stages: \emph{i)} workspace perception: in which an overhead camera captures the physical task context and detects the user's pointing gesture, \emph{ii)} context-aware content generation: in which a multimodal AI model interprets the captured scene and produces task-relevant guidance 
and \emph{iii)} robot-mediated projection: in which the robotic arm projects the generated content alongside the user's physical task. 

Through the different interaction modes (i.e., math homework, generate image, and draw), we demonstrate a range of desk-based learning and creative activities. Our work investigates whether embedding AI guidance directly in the physical workspace through AIfred improves short term learning transfer, creative output quality, and user experience compared to conventional screen-based AI assistance. We additionally measure context-switching behavior and task engagement to understand the mechanisms through which spatial co-location may produce these effects.

To evaluate whether AIfred offers measurable benefits over screen-based alternatives, we conducted a user study comparing AIfred against ChatGPT (GPT-5.6 Luna) running on a laptop. Participants ($n=36$) were tasked to complete math-related assignments, image-generation from sketches, and drawing tasks (i.e., the different interaction modes) using both approaches. We measured outcome quality, short-term learning transfer, task completion time, and observed physical-digital context switches. Participants also rated their subjective experience such as perceived learning support, cognitive demand, perceived context switching, and perceived productivity (see Sec.~\ref{subsec:user_study}).

Our results show that AIfred's benefits concentrate in short-term learning transfer and drawing quality. In the math assignment, the two conditions performed comparably while assistance was available (6.7/10 with AIfred vs. 7.3/10 with ChatGPT, $p = .41$), but once assistance was removed the ChatGPT group fell to 4.4/10 while the AIfred group held at 7.0/10, a 60\% higher short-term transfer score ($p = .003$). In the drawing task, three independent art and design professors ranked drawings produced with AIfred best in 92\% of cases, with qualitative comparisons showing gains in overall form, proportions, and line quality. Participants also reported higher perceived learning support ($p < .001$) and higher cognitive demand ($p = .03$) with AIfred, with no accompanying loss in perceived productivity ($p = .61$). We hope these findings inspire future research on embodied AI systems that integrate AI assistance directly into physical learning and creative activities.

\section{Related Work}
\label{sec:soa}

Robots are increasingly entering educational settings as tutors, companions, and collaborative learners. The physical presence of robots in learning environments has been shown to influence student engagement, motivation, and learning outcomes in ways distinct from purely screen-based systems~\cite{kennedy2016social, belpaeme2018social}. Educational robots span a range of embodiments, from humanoid social robots like NAO~\cite{belpaeme2013multimodal, castello2018robochain}, Jibo~\cite{scassellati2018improving} and Pepper~\cite{pandey2018pepper}, to zoomorphic companions~\cite{kory2013storytelling} and desktop robots used for support in specific learning activities~\cite{jung2017engaging}. Much of this research focuses on the robot as a tutor that provides feedback, demonstrates empathy, adapts its tutoring strategies, or serves as a peer learner~\cite{saerbeck2010expressive,leyzberg2018robotutor}. In this realm, studies show that robots capable of expressing social cues through gaze, gesture, and movement can provoke higher engagement than screen-based agents~\cite{kennedy2016social,leite2013social}. This social dimension is particularly important for long-term interaction, where repeated exposure may reduce novelty effects and require more sophisticated behavioral design~\cite{kanda2007two}.

Beyond tutoring, recent works explore robot manipulators as demonstrators in STEM education. Projects like CoWriter~\cite{hood2015children} uses robots to apply the learning-by-teaching paradigm, where children correct a ``poorly writing'' robot, thereby reflecting on their own letter formation. Other work investigates how robots can physically demonstrate scientific principles, assist in assembly tasks, or act as tangible programming interfaces~\cite{alves2019exploring}. These systems illustrate how embodiment can move beyond social expression toward functional support that leverages the robot's physical capabilities. 
However, most educational robots have been studied primarily as tutors or learning companions that provide feedback, explanations, or social interaction, rather than as systems that augment the learner's ongoing physical work~\cite{belpaeme2018social, kennedy2016social}. While prior research has explored physical-material perception and projection-based assistance, these capabilities have generally been investigated separately from adaptive, AI-driven interpretation of the learner's workspace \cite{wellner1993digitaldesk, zhou2026projectasemihumanoidroboticteaching}. In addition, previous works require users to shift attention to the robot as a separate entity. This interaction model positions the robot as a tutor standing beside the user or as a companion sharing the space, rather than as an integrated assistant embedded within the users workspace~\cite{educsci13040347}.

The integration of digital information into physical environments has been a central theme in human-computer interaction research for over three decades. SAR extends augmented reality by embedding digital content directly into the environment through projection~\cite{bimber2005spatial}. DigitalDesk pioneered the concept of a projector-camera desk that could recognize physical documents, augment them with digital content, and respond to finger-based interaction~\cite{wellner1993digitaldesk}. Subsequent systems such as metaDESK~\cite{ishii1997tangible} and Urp~\cite{underkoffler1999urp} extended this idea by integrating tangible user interfaces that users could manipulate while digital projections updated in real time. These systems illustrated how spatial computing can allow users to think and act with physical objects while digital information remained co-located on the same surface.

More recent work has explored robot-mediated projection system that can dynamically reposition digital content across a workspace. LuminAR~\cite{linder2010luminar} and ELEGNT~\cite{hu2025elegnt} represent the two systems whose design is most similar to ours (AIfred). LuminAR introduced a robotic desk lamp equipped with a projector and camera, enabling projection and touch interaction on arbitrary surfaces within reach. The system allowed users to move the lamp to project content wherever needed, blending the mobility of a physical lamp with interactive projection capabilities. Building on this lamp-like form factor, ELEGNT integrated a 6-DOF robotic arm with a projector, camera, and light to explore how expressive movement design can enhance human-robot interaction in domestic settings. While both LuminAR and ELEGNT demonstrate the potential of robotic projection platforms, LuminAR focuses on manual repositioning for interactive projection, and ELEGNT focuses on expressive movement as a communication channel and explores a broad design space for home interaction. Neither system addresses real-time task interpretation through AI or embeds learning support into ongoing desk work.

\begin{figure*}[!t]
    \centering

    \begin{minipage}[t]{0.318\textwidth}
        \centering
        \includegraphics[width=\linewidth]{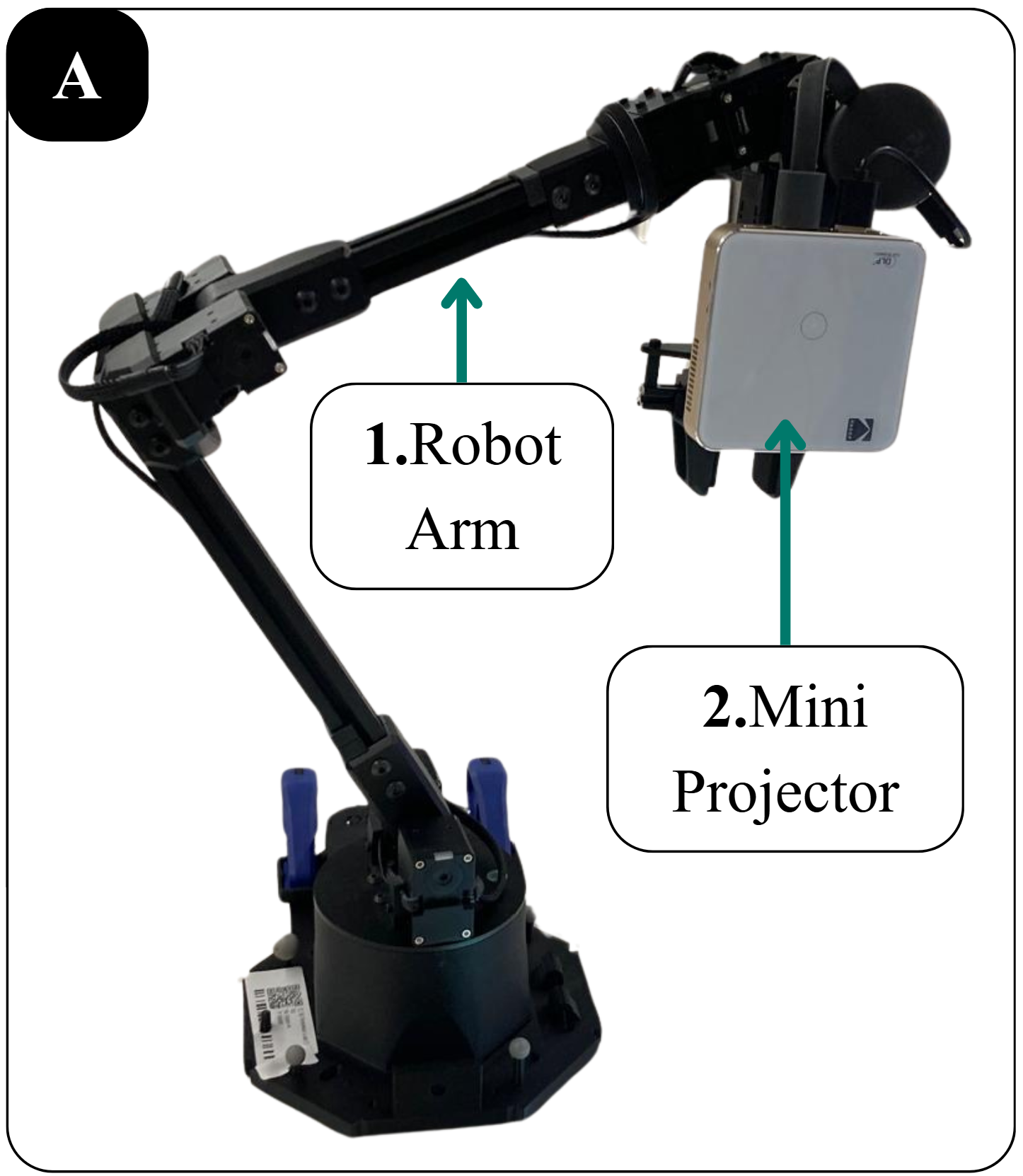}
    \end{minipage}
    \hspace{0.01\textwidth}
    \begin{minipage}[t]{0.658\textwidth}
        \centering
        \includegraphics[width=\linewidth]{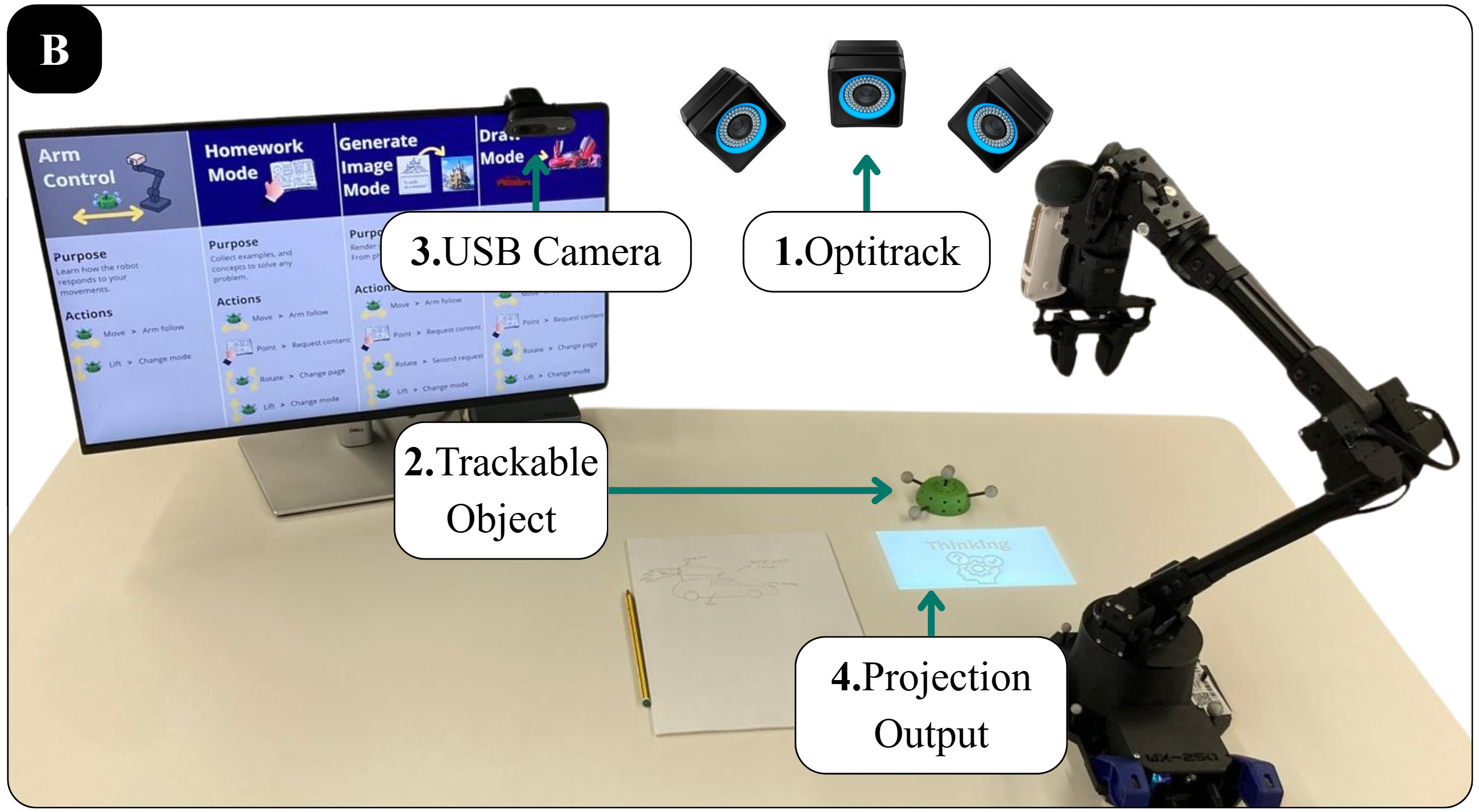}
    \end{minipage}

    \vspace{0.3em}
    \caption{AIfred prototype (A) consists of a robotic arm (A.1) with a projector mounted on the end-effector (A.2). The experimental setup (B) includes a motion capture system (B.1), a physical trackable object to interact with the arm (B.2), an overhead camera (B.3) and the respective context-relevant generated content projection output (B.4).}
    \label{fig:AIfred-system-overview}
\end{figure*}
\vspace{-3pt}

Projection-based interfaces have also been applied specifically to learning contexts. Systems such as Augmented Studio and Assemblé project instructional overlays onto physical materials to support craft, assembly, and making tasks ~\cite{zhu2019augmented,li2019reactile}. These systems demonstrate that on-site projection can reduce the cognitive cost of switching between instructions and action. However, they typically rely on predefined content, pre-registered materials, or scripted tutorials, and do not interpret or adapt to evolving work. AIfred, in contrast, generates task-relevant guidance through generative AI in real time, directly augmenting the user's outcomes, performance, and learning as the task unfolds.

\section{Methodology}
AIfred's prototype~\footnote{All code and materials used in this work is publicly at: \href{https://github.com/IERoboticsAILab/AIfred}{https://github.com/IERoboticsAILab/AIfred}.} (Fig.~\ref{fig:AIfred-system-overview} A) is composed of a robotic arm (A.1) with a mini projector attached at its end-effector (A.2), which is integrated in the user workspace (Fig.~\ref{fig:AIfred-system-overview} B). The robotic arm is placed at the edge of the desk so that its end-effector can reach and orient toward the user’s desk. To know where information should be projected, the system uses an OptiTrack motion-capture system (B.1) to track the pose of the robot base and a trackable object placed on the desk (B.2). An inverse kinematics solver then computes the arm configuration required to place the mounted projector toward the target region while maintaining a fixed projection height over the desk. The tracked object allows the user to reposition the projected content to wherever it is most relevant on the desk. 
Complementarily, an overhead camera (B.3) provides a top-down view of the workspace and captures the physical task context, along with the user’s pointing gesture. These visual inputs are used by the software stack to infer the user’s task context and generate the corresponding projected output (B.4).

\begin{figure*}[!t]
    \centering
    \includegraphics[width=0.98\textwidth]{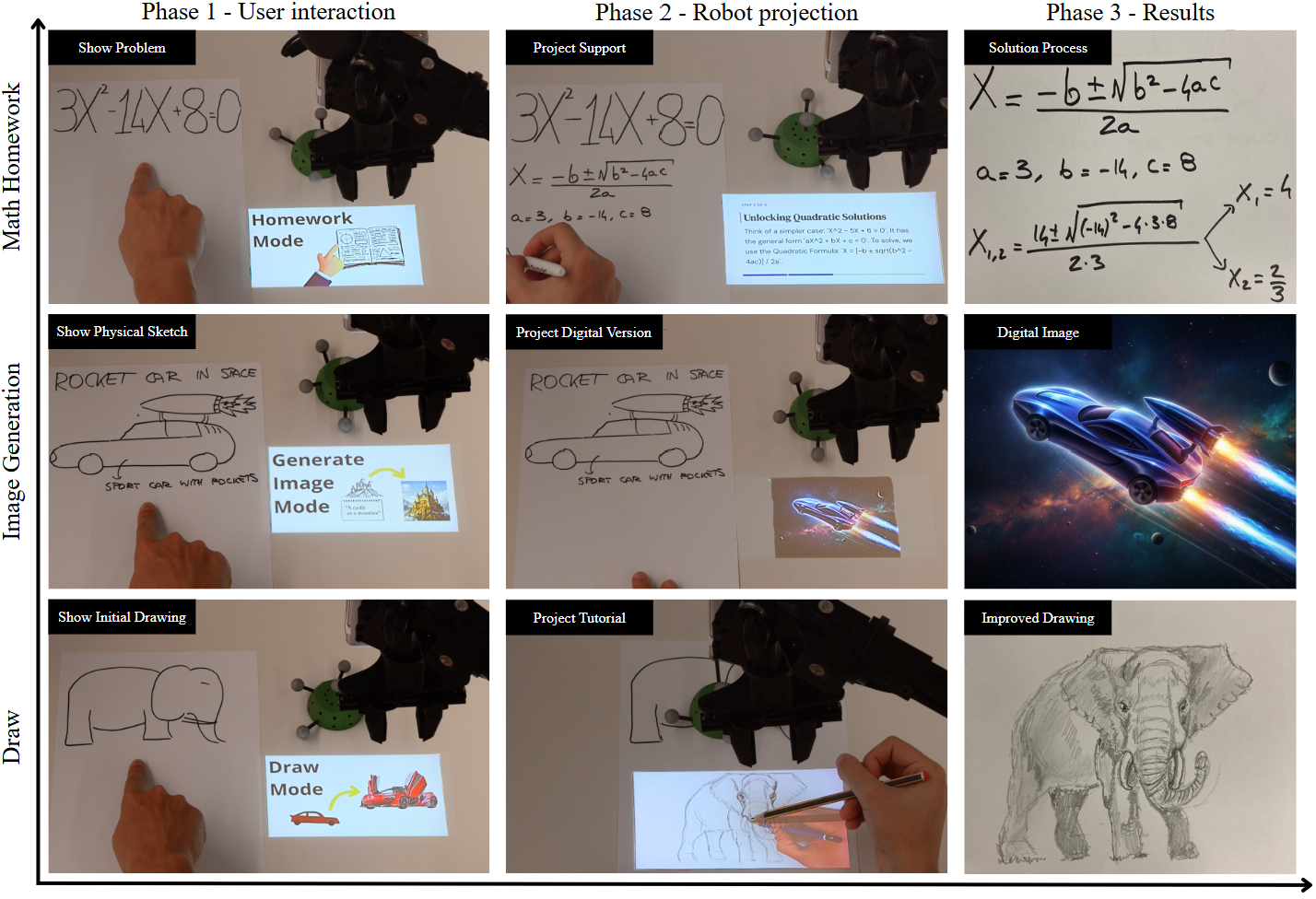}
    \caption{Illustration of AIfred interaction scenarios, organized by interaction mode (math homework, generate image, and draw) and interaction phase: Phase 1, user interaction (request assistance for task); Phase 2, content generation and robot projection (perform task); and Phase 3, result (task done).}
    \label{fig:AIfred-interaction-modes-phases}
\end{figure*}

\subsection{Spatial assistance pipeline}
\label{subsec:software}
The spatial assistance pipeline of AIfred combines three stages: \emph{i)} workspace perception, \emph{ii)} context-aware content generation and \emph{iii)} robot-mediated projection. During the workspace perception, AIfred observes the user's desk through the overhead camera, capturing the physical materials involved in the current task (e.g., sketches, printed content, handwritten notes). The system continuously monitors hand gestures, using MediaPipe library, to identify the region of the workspace for which assistance is being requested. When a pointing gesture is detected, it triggers a screenshot of the relevant area. Next, in the context-aware content generation, the captured image is passed to a multimodal AI model, powered by Google's Gemini API, together with the currently active interaction mode. We use \textit{gemini-3.8-flash} for scene understanding and text-based reasoning, and \textit{gemini-2.5-flash-image} for image generation. The interaction mode constrains how the scene is interpreted and what type of assistance is generated. The model interprets the user's current work, and through dedicated structured prompts that encodes AIfred's tutoring approach, it generates task-relevant content. Finally, in the robot-mediated projection, the generated content is projected onto the desk surface at the location indicated by the tracked object. 

\subsection{Interaction Modes}
\label{subsec:interaction_modes}

To evaluate AIfred, we programmed three interaction modes that represent different forms of desk activity. These modes were selected to cover a progression from AI-assisted reasoning, to creative transformation, and drawing quality. In all modes, the user remains seated at the desk and interacts with physical materials, while AIfred provides spatially situated visual support through the robot-mounted projector. The modes differ in the type of user activity, the role of AI assistance, and the way projected content supports the task. Fig.~\ref{fig:AIfred-interaction-modes-phases} illustrates the three interaction modes.

\subsubsection{Math homework mode}
Math homework mode instantiates AIfred as a learning assistant. In this mode, the user works on a written or printed exercise on the desk and requests assistance by pointing to the relevant material (\emph{i}). The overhead camera captures the selected area, and the multimodal model interprets the problem. AIfred then projects learning support relevant information next to the physical work material (\emph{ii}). The design goal is to provide scaffolding rather than answer replacement. Instead of directly solving the task for the user, the system generates hints, relevant formulas, intermediate reasoning cues, or analogous examples that assist the user to continue the solution process (\emph{iii}). 

\subsubsection{Generate image mode}
Generate image mode explores AIfred as a bridge between physical sketching and digital content generation. In this mode, the user produces a sketch, diagram, or visual idea on the desk, and AIfred interprets the selected region as input for AI-supported visual transformation (\emph{i}). The system can generate a refined image and digitally cleaned version of the user’s physical sketch, which is then projected back on to the desk (\emph{ii}). The result digital image is then saved in the computer (\emph{iii}).

\subsubsection{Draw mode} 
Draw mode instantiates AIfred as a spatial guide for drawing on paper. In this mode, the user first attempts to draw a figure directly on paper without any assistance. After this initial attempt, the user points to the initial drawing, and the overhead camera captures the selected workspace region (\emph{i}). The multimodal AI model interprets the user’s drawing and identifies what type of visual support may assist the user improve it. AIfred then projects visual examples (e.g., reference images, drawing tutorials, videos) directly onto the user’s paper (\emph{ii}). This allows the user to compare reference material and their own work without shifting attention between screen and desk. The user can follow the instruction on the desk in order to improve the initial drawing (\emph{iii}).

\section{User study}
\label{subsec:user_study}
We used a mixed experimental design to compare AIfred with ChatGPT, which we
selected as the screen-based baseline because it is the AI tool students most commonly use for academic work~\cite{mostusedtool}. Participants were assigned to one of two conditions: one group completed the tasks using ChatGPT on a laptop, while the other completed the same tasks using AIfred. This design allowed us to compare the two systems while keeping task categories consistent across conditions.

We designed three main desk-based task categories: \textit{math assignment}, \textit{image generation}, and \textit{drawing}. In the math-assignment task, participants practiced problem solving on a quadratic equation, selecting from a pool of five equations of comparable difficulty to ensure equivalence across participants and conditions. In the image-generation task, participants practice rendering a physical sketch (any visual idea they drew on a paper) into a digital image. In the drawing task, users practice drawing a chosen figure among a car, a motorbike, a lion or an elephant.

The experiment began with participants performing a math assignment and an image generation task with the assistance of the assigned system (AIfred or ChatGPT). Participants then performed three drawing tasks in sequence: the first one with no assistance, the second with the assistance of their assigned system and the third with the assistance of the opposite system, as part of a crossover trial. This means that participants in the ChatGPT condition used AIfred for the third drawing task, while participants in the AIfred condition used ChatGPT. This crossover was included to enable direct within-participant comparisons of drawing outcomes across systems. Finally, participants completed an additional math assignment, solving another quadratic-equation exercise without any assistance. This task was included to assess short-term learning transfer. This metric allows us to examine whether participants could still apply relevant reasoning after the assistance was removed, with a time distance between the two math assignment of on average thirty-five minutes.

\subsection{Performance metrics}
\label{subsec:metrics}
We collected both quantitative and qualitative measures to evaluate how AIfred affected participants’ behavior, task performance, and user experience during desk-based activities compared with standard ChatGPT (GPT-5.6 Luna). Quantitative measures were divided into behavioral and performance measures. 

Behavioral measures focused on physical-digital context switching, where context switch was defined as a post-experiment human supervision of the videos and count of visual attention shifts between the physical desk and the laptop screen. Performance measures captured task completion time, outcome quality, and improvement percentages such as short-term learning transfer.

Outcome quality was evaluated using task-specific criteria and grading procedures, tailored to the nature of each task. For the drawing task, outcome quality was assessed by three independent art and design professors that independently ranked each participant's three drawings from best (1st) to worst (3rd). Inter-rater agreement was quantified using Kendall's coefficient of concordance ($W=.86$), indicating that the three professors were not guessing or disagreeing on the rankings but largely converged on the same ordering for each participant.

For the math-assignment task, solutions were graded by four independent Google Gemini Flash 3.6 grading agents, applying the same fixed rubric. To avoid condition-dependent bias, each agent received the  participants' solutions anonymised and in random order, so that the condition (AIfred, ChatGPT, or no assistance) could not be inferred. The rubric decomposed each quadratic-equation solution into eight measurable criteria: problem setup, method selection, discriminant calculation, algebraic working, first root, second root, simplification, and verification, yielding a final grade on a 1--10 scale. Across all solutions, the mean inter-agent standard deviation was 0.53 points on this scale, indicating that the fixed rubric produced consistent and reproducible grades across independent agents.

Self-reported user-experience measures were collected using pre and post experiment questionnaires. In the post-experiment questionnaire, participants also evaluated their experience across six dimensions, rated on a Likert scale from 1 to 5. Perceived learning support captured whether the system supported understanding, engagement, and motivation throughout the learning process. Cognitive demand captured the mental effort participants felt they invested in completing the tasks, as well as clarity of the stated goals and procedures. Perceived productivity captured whether the system saved time and enabled participants to accomplish more than they otherwise could. Innovation captured whether the system helped participants generate new ideas and engage with the task in creative, non-traditional ways. User satisfaction captured whether the system met participants' needs and improved their overall satisfaction with the task. Finally, perceived context-switching disruption captured whether shifting attention between the physical workspace and the digital source interrupted concentration, for this measure alone, the scale was inverted, with 1 indicating no perceived disruption and 5 indicating high disruption.

We recruited 36 participants ($n=36$) across the university campus. 17 were men and 19 were women. The participants age ranges from 19 to 50. Participants represented a diverse range of academic backgrounds, spanning humanistic, social-science, business, and engineering. They were generally acquainted with technologies such as robotics and AI, reporting average familiarity scores of 5.8/10 and 7.0/10 respectively.

\section{Results}

\subsection{Drawing Quality}
\begin{figure}[htbp]
    \centering
    \hspace*{-0.25cm}
    \includegraphics[height=4cm]{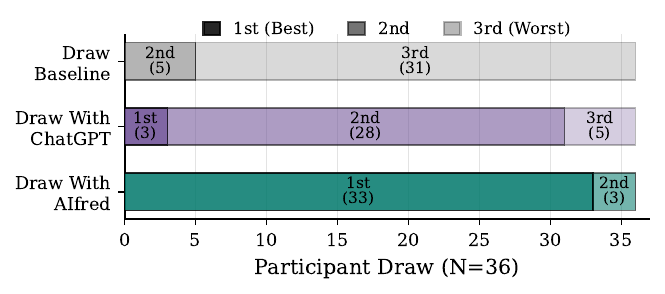}
    \caption{Rank of each participant's three drawing from from best (1st) to worst (3rd).}
    \label{fig:draw_results}
\end{figure}

\begin{figure}[htbp]
    \centering
    \includegraphics[width=0.9\columnwidth]{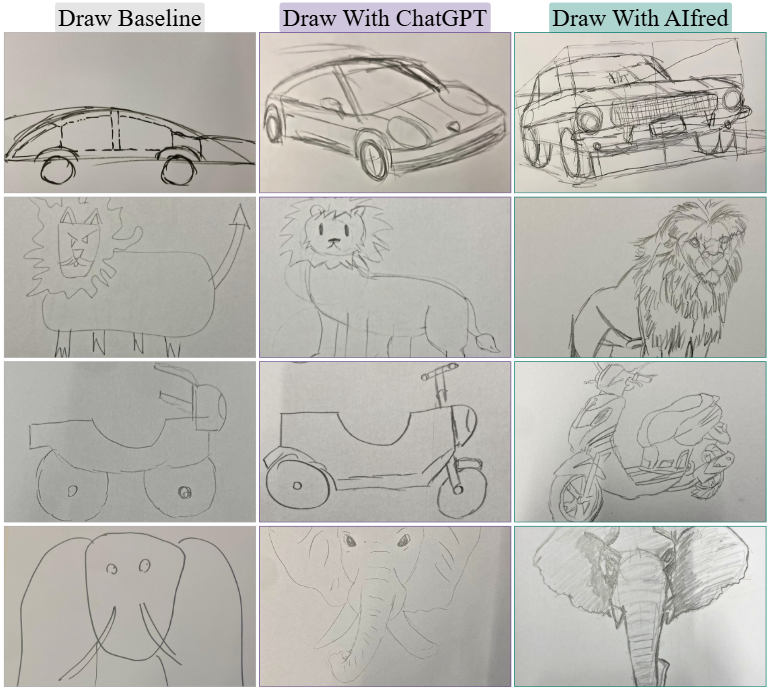}
    \caption{Qualitative comparison between output drawing of few participants without help (left), with ChatGPT help (center), and with AIfred help (right).}
    \label{fig:DrawOutcomes}
\vspace{-3pt}
\end{figure}

Fig.~\ref{fig:draw_results} summarizes the art and design professor rankings of the drawing task. Each participant produced three drawings (baseline without assistance, with ChatGPT, and with AIfred), and each of the three professors ranked them from 1st (best) to 3rd (worst). Fig.~\ref{fig:draw_results} reports the median of the three ranks per drawing, and each bar shows, for one condition, how many of the 36 participants had their drawing placed in each rank position. Baseline drawings were ranked 3rd in 31 of 36 cases (86\%) and never 1st. Drawings made with ChatGPT were ranked 2nd in 28 cases (78\%), 1st in only 3, and 3rd in 5. Drawings made with AIfred were ranked 1st in 33 of 36 cases (92\%), 2nd in the remaining 3, and never 3rd. A Friedman test confirmed that the three conditions differ overall ($\chi^2 = 57.06$, $p < .0001$), and pairwise Wilcoxon signed-rank tests confirmed that every pairwise ordering is individually significant (all $p < .00001$, except baseline vs.\ ChatGPT, $p = .00002$). Fig.~\ref{fig:DrawOutcomes} shows representative drawings produced without assistance, with ChatGPT, and with AIfred. The drawings produced with AIfred show improvements in overall form and in finer details, including proportions and line quality.

\subsection{Math Assignment and Short-Term Learning Transfer}
\begin{figure}[htbp]
    \centering
    \includegraphics[height=6cm]{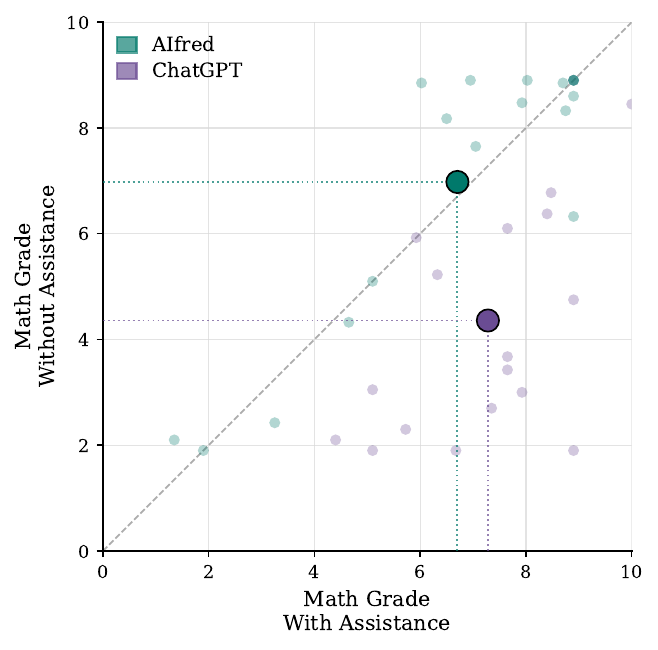}
    \caption{Quantitative results for the math assignment task. Final grade with ChatGPT or AIfred assistance versus final grade later when assistance was removed.}
    \label{fig:homework_results}
    \vspace{-5pt}
\end{figure}

Fig.~\ref{fig:homework_results} summarizes performance results for the math assignment task. The figure shows participants' scores with assistance against their scores on a second problem without assistance. When supported by the assigned system, participants’ task scores do not differ significantly between AIfred and ChatGPT (6.7/10 vs. 7.3/10; $p=.41$). But once the assistance was removed, the mean score of participants previously assisted by ChatGPT falls to 4.4/10, a 40\% decrease. In contrast, participants previously assisted by AIfred maintain their performance, scoring 7.0/10 without assistance, a 4.5\% increase, representing a 60\% higher short-term learning transfer score compared to participants previously assisted by ChatGPT ($p=.003$).

\subsection{Image Generation Task Performance}
\begin{figure}[htbp]
    \centering
    \includegraphics[height=5.5cm]{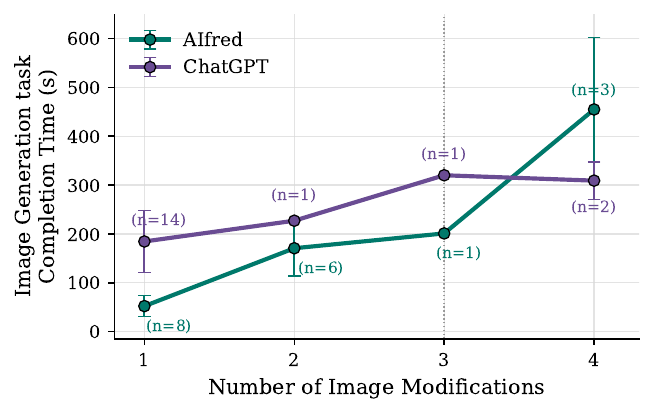}
    \caption{Mean image-generation task completion time as a function of the number of time the output image was modified for AIfred and ChatGPT.}
    \label{fig:generate_image_results}
\end{figure}

Fig.~\ref{fig:generate_image_results} reports the completion time with respect to the number of image modifications during the image generation task. In the figure, markers represent mean completion times, error bars show variability, and annotations indicate the number of observations at each modifications count. For one, two, and three modifications, AIfred shows lower completion times than ChatGPT. ChatGPT, however, becomes faster at four modifications. AIfred accelerates the initial modification by allowing users to point directly to their physical sketch, avoiding image capture and upload. However, each subsequent image modification requires users to physically modify and re-render the sketch, whereas ChatGPT supports faster revisions through text prompts. On average, task completion time with AIfred is 167 seconds and with ChatGPT it is 208 seconds ($p=.32$).

\subsection{Physical-Digital Context-Switching Behavior}
\begin{figure}[htbp]
    \centering
    \hspace*{-0.28cm}
    \includegraphics[height=5.5cm]{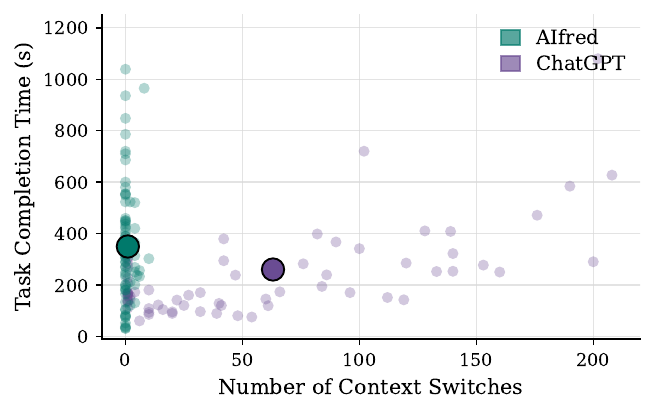}
    \label{fig:context_switching_quantitative}
    \caption{Observed physical-digital context switches and task-completion times comparing AIfred and ChatGPT.}
    \label{fig:context_switching}
\end{figure}

Fig.~\ref{fig:context_switching} compares task completion time across all tasks with respect to measured context switches between the physical workspace and digital screen. Transparent points represent individual trials, and outlined markers show condition means. Participants performed on average 1 switch with AIfred and 63 with ChatGPT, a 98\% reduction. Participants using AIfred looked at the screen, where we displayed interaction instructions, only once per task on average. Mean task completion time was higher with AIfred (349.5 seconds vs 259.9 seconds; $p=.02$). The additional time reflects sustained physical engagement with the task materials. In learning contexts, longer time-on-task can be associated with deeper processing and improved retention (see Discussion, Sec.~\ref{sec:discussion}).

\subsection{User Experience}
\begin{figure}[!htbp]
    \centering
    \includegraphics[height=4.7cm]{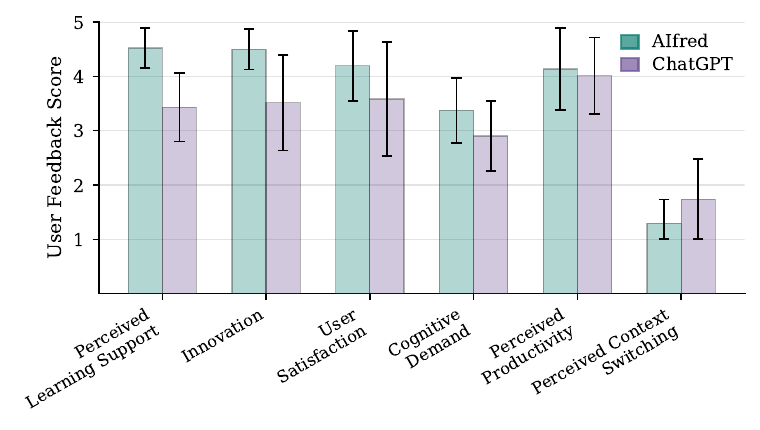}
    \caption{Self-reported user-experience ratings across different metrics comparing AIfred and ChatGPT users.}
    \label{fig:qualitative}
\end{figure}

Fig.~\ref{fig:qualitative} summarizes six user-experience dimensions explored in this paper. AIfred scored higher than ChatGPT in perceived learning support (4.5/5 vs. 3.4/5; $p<.001$), the metric most directly aligned with the system's educational goal. Participants also rated AIfred higher in innovation (4.5/5 vs. 3.5/5; $p<.001$) and user satisfaction (4.2/5 vs. 3.5/5; $p=.04$). Cognitive demand trended higher with AIfred (3.4/5 vs. 2.9/5; $p=.03$), consistent with the greater task engagement observed in the behavioral data. Perceived productivity was statistically indistinguishable across conditions (4.1/5 vs. 4.0/5; $p=.61$), indicating that the longer completion times with AIfred did not translate into a subjective sense of inefficiency. Finally, both groups reported low perceived context switching (1.3/5 vs. 1.7/5; $p=.03$), despite the 98\% difference in observed switches.

\section{Discussion}
\label{sec:discussion}

In this paper, we investigated whether delivering AI assistance directly into the physical workspace changes how users learn and create during desk-based activities. Both behavioral and self-reported findings show that AIfred improves the character of the interaction compared to ChatGPT. Participants working with AIfred sustained their engagement on paper, produced drawings that art and design professors preferred in the large majority of cases, and retained more of what they had practiced once the assistance was withdrawn. We suspect screen-based guidance often leads to passive transcription, as users can copy content with minimal reformulation. By contrast, projected guidance requires users to actively reinterpret and reproduce information by hand, promoting deeper task engagement.

The results also reveal that the benefit of spatial co-location varies across tasks. For activities such as drawing, where the user must continuously compare a reference against the proportions, angles and placement of their own lines, projected assistance was preferred. For activities such as solving a quadratic equation, AIfred and ChatGPT performed comparably while assistance was available ($6.7/10$ vs.\ $7.3/10$, $p = .41$). This means that the task types matter. Symbolic content can be read from a screen, held briefly in memory, and reproduced on paper. In contrast, spatial guidance relies heavily on direct visual alignment with the physical work. Every context switch to a screen breaks this visual reference, forcing the user to mentally re-align the guidance onto their paper upon looking back. As a result, co-locating guidance matters most when the instructions and the task share a unified spatial frame. Our findings imply that robot-mediated projection is not a general-purpose replacement for screen-based assistance, but a targeted one. 

Finally, participants using ChatGPT performed 63 observed physical-digital context switches, compared with 1 with AIfred, yet both groups rated the disruption as low ($1.7/5$ and $1.3/5$). This indicates that participants did not consciously register screen-to-paper switching as disruptive, even though it occurred frequently and coincided with less favorable outcomes, such as lower short-term transfer and lower drawing quality. We interpret this as evidence that attentional fragmentation can operate below the threshold of subjective awareness. Users may be sufficiently accustomed to alternating between physical and digital media that the cost of switching goes unnoticed, even as it quietly shapes performance.

\subsection{Limitations and Future Work}
The current prototype relies on an OptiTrack motion-capture system for spatial tracking, which, while suitable for fast prototyping, limits deployment outside controlled lab settings and can be replaced by vision-based tracking. The fast prototyping choice reflects the objective of the current study, which is not to optimize robot interaction design, but to evaluate whether robot-mediated projection can spatially embed AI assistance into physical desk-based learning, improving user experience and performance. Additionally, AIfred's current design prioritizes functional projection over expressive behavior. Future work should explore expressive, movement-driven robot behavior, to further enhance user engagement.

\section{Conclusions}
In this paper, we present AIfred, a system for delivering generative AI assistance directly within the user's physical workspace. We use a desk-based robotic arm with a projector mounted on the end-effector to develop three interaction modes: math homework, image generation, and drawing. We conduct a user study with 36 participants to compare AIfred against a screen-based AI assistant across these desk-based tasks. Our results indicate that spatially co-located assistance improves drawing quality, with art and design professors ranking AIfred drawings first in 33 of 36 cases. Short-term learning transfer was also improved, with participants scoring 2.6 points higher on a 10-point scale. Behavioral analysis further shows that participants performed 63 physical-digital context switches with the screen-based assistant and 1 with AIfred, yet rarely registered this switching as disruptive, suggesting that the cost of attentional fragmentation is largely invisible. Our findings position systems such as AIfred as a promising direction for AI-assisted learning tools that support understanding.

\section*{Acknowledgments}
AI was used during the preparation of this manuscript for grammar and language editing, literature research, code assistance, and data visualization/plot generation. All research decisions, analyses, and final content were reviewed and verified by the authors.

\bibliography{references}

@article{pandey2018pepper,
  title={Pepper: The first machine of its kind},
  author={Pandey, Amit Kumar and Gelin, Rodolphe and Robot, AMPSH},
  journal={IEEE Robotics \& Automation Magazine},
  volume={25},
  number={3},
  pages={40--48},
  year={2018}
}

@article{
belpaeme2018social,
author = {Tony Belpaeme  and James Kennedy  and Aditi Ramachandran  and Brian Scassellati  and Fumihide Tanaka },
title = {Social robots for education: A review},
journal = {Science Robotics},
volume = {3},
number = {21},
pages = {eaat5954},
year = {2018},
doi = {10.1126/scirobotics.aat5954},
URL = {https://www.science.org/doi/abs/10.1126/scirobotics.aat5954},
eprint = {https://www.science.org/doi/pdf/10.1126/scirobotics.aat5954}}

@misc{zhou2026projectasemihumanoidroboticteaching,
      title={ProjecTA: A Semi-Humanoid Robotic Teaching Assistant with In-Situ Projection for Guided Tours}, 
      author={Hanqing Zhou and Yichuan Zhang and Zihan Zhang and Wei Zhang and Chao Wang and Pengcheng An},
      year={2026},
      eprint={2601.11328},
      archivePrefix={arXiv},
      primaryClass={cs.HC},
      url={https://arxiv.org/abs/2601.11328}, 
}

@Article{educsci13040347,
AUTHOR = {Hiroi, Yutaka and Ito, Akinori},
TITLE = {A Robotic System for Remote Teaching of Technical Drawing},
JOURNAL = {Education Sciences},
VOLUME = {13},
YEAR = {2023},
NUMBER = {4},
ARTICLE-NUMBER = {347},
URL = {https://www.mdpi.com/2227-7102/13/4/347},
ISSN = {2227-7102},
DOI = {10.3390/educsci13040347}
}

@article{mostusedtool,
author = {S N, Archana and V R, Renjith and P K, Padmakumar and Cheruvil, Shajitha and Aboobaker, Nimitha},
year = {2025},
month = {10},
pages = {},
title = {AI assisted learning and research: an exploratory study among university students and scholars},
volume = {4},
journal = {Discover Education},
doi = {10.1007/s44217-025-00814-x}
}

@book{OECD2026PISA,
  author    = {{OECD}},
  title     = {PISA 2025 Results (Volume I): Future-Ready Students},
  year      = {2026},
  publisher = {OECD Publishing},
  address   = {Paris},
  series    = {PISA},
  doi       = {10.1787/73451bc5-en},
  url       = {https://doi.org/10.1787/73451bc5-en}
}

@article{mueller2014pen,
  author  = {Mueller, Pam A. and Oppenheimer, Daniel M.},
  title   = {The Pen Is Mightier Than the Keyboard: Advantages of Longhand Over Laptop Note Taking},
  journal = {Psychological Science},
  volume  = {25},
  number  = {6},
  pages   = {1159--1168},
  year    = {2014},
  doi     = {10.1177/0956797614524581}
}

@misc{hu2025elegnt,
  author       = {Hu, Yuhan and Huang, Peide and Sivapurapu, Mouli and Zhang, Jian},
  title        = {{ELEGNT}: Expressive and Functional Movement Design for Non-anthropomorphic Robot},
  year         = {2025},
  eprint       = {2501.12493},
  archivePrefix = {arXiv},
  primaryClass = {cs.RO},
  doi          = {10.48550/arXiv.2501.12493}
}

@book{bimber2005spatial,
  author    = {Bimber, Oliver and Raskar, Ramesh},
  title     = {Spatial Augmented Reality: Merging Real and Virtual Worlds},
  publisher = {A K Peters/CRC Press},
  address   = {Wellesley, MA, USA},
  year      = {2005},
  isbn      = {9781568812304},
  doi       = {10.1201/b10624}
}

@inproceedings{linder2010luminar,
  author    = {Linder, Natan and Maes, Pattie},
  title     = {LuminAR: Portable Robotic Augmented Reality Interface Design and Prototype},
  booktitle = {Adjunct Proceedings of the 23nd Annual ACM Symposium on User Interface Software and Technology},
  series    = {UIST '10},
  pages     = {395--396},
  year      = {2010},
  publisher = {Association for Computing Machinery},
  address   = {New York, NY, USA},
  doi       = {10.1145/1866218.1866237}
}

@inproceedings{ishii1997tangible,
  author    = {Ishii, Hiroshi and Ullmer, Brygg},
  title     = {Tangible Bits: Towards Seamless Interfaces Between People, Bits and Atoms},
  booktitle = {Proceedings of the SIGCHI Conference on Human Factors in Computing Systems},
  series    = {CHI '97},
  pages     = {234--241},
  year      = {1997},
  publisher = {Association for Computing Machinery},
  address   = {New York, NY, USA},
  doi       = {10.1145/258549.258715},
  isbn      = {0897918029}
}

@article{wellner1993digitaldesk,
  author  = {Wellner, Pierre},
  title   = {Interacting with Paper on the DigitalDesk},
  journal = {Communications of the ACM},
  volume  = {36},
  number  = {7},
  pages   = {87--96},
  year    = {1993},
  doi     = {10.1145/159544.159630}
}

@inproceedings{ullmer1997metadesk,
  author    = {Ullmer, Brygg and Ishii, Hiroshi},
  title     = {The metaDESK: Models and Prototypes for Tangible User Interfaces},
  booktitle = {Proceedings of the 10th Annual ACM Symposium on User Interface Software and Technology},
  series    = {UIST '97},
  pages     = {223--232},
  year      = {1997},
  publisher = {Association for Computing Machinery},
  address   = {New York, NY, USA},
  doi       = {10.1145/263407.263551},
  isbn      = {0897918819}
}

@inproceedings{underkoffler1999urp,
  author    = {Underkoffler, John and Ishii, Hiroshi},
  title     = {Urp: A Luminous-Tangible Workbench for Urban Planning and Design},
  booktitle = {Proceedings of the SIGCHI Conference on Human Factors in Computing Systems},
  series    = {CHI '99},
  pages     = {386--393},
  year      = {1999},
  publisher = {Association for Computing Machinery},
  address   = {New York, NY, USA},
  doi       = {10.1145/302979.303114},
  isbn      = {0201485591}
}

@article{kennedy2016social,
  author  = {Kennedy, James and Baxter, Paul and Senft, Emmanuel and Belpaeme, Tony},
  title   = {Social Robot Tutoring for Child Second Language Learning},
  journal = {ACM/IEEE International Conference on Human-Robot Interaction},
  pages   = {231--238},
  year    = {2016},
  publisher = {IEEE Press},
  doi     = {10.1109/HRI.2016.7451757}
}

@inproceedings{belpaeme2013multimodal,
  author    = {Belpaeme, Tony and Baxter, Paul E. and Read, Robin and Wood, Rachel and Cuay{\'a}huitl, Heriberto and Kiefer, Bernd and Racioppa, Stefania and Kruijff-Korbayov{\'a}, Ivana and Athanasopoulos, Georgios and Enescu, Valentin and Looije, Rosemarijn and Neerincx, Mark and Demiris, Yiannis and Ros-Espinoza, Raquel and Beck, Aryel and Ca{\~n}amero, Lola and Hiolle, Antoine and Lewis, Matthew and Baroni, Ilaria and Nalin, Marco and Cosi, Piero and Paci, Giulio and Tesser, Fabio and Sommavilla, Giacomo and Humbert, Remi},
  title     = {Multimodal Child-Robot Interaction: Building Social Bonds},
  booktitle = {Journal of Human-Robot Interaction},
  volume    = {1},
  number    = {2},
  pages     = {33--53},
  year      = {2013},
  doi       = {10.5898/JHRI.1.2.Belpaeme}
}

@article{scassellati2018improving,
  author  = {Scassellati, Brian and Boccanfuso, Laura and Huang, Chien-Ming and Mademtzi, Marilena and Qin, Meiying and Salomons, Nicole and Ventola, Pamela and Shic, Frederick},
  title   = {Improving Social Skills in Children with ASD Using a Long-Term, In-Home Social Robot},
  journal = {Science Robotics},
  volume  = {3},
  number  = {21},
  pages   = {eaat7544},
  year    = {2018},
  doi     = {10.1126/scirobotics.aat7544}
}

@inproceedings{kory2013storytelling,
  author    = {Kory, Jacqueline and Breazeal, Cynthia},
  title     = {Storytelling With Robots: Learning Companions for Preschool Children's Language Development},
  booktitle = {2013 IEEE International Symposium on Robot and Human Interactive Communication (RO-MAN)},
  pages     = {643--648},
  year      = {2013},
  publisher = {IEEE},
  doi       = {10.1109/ROMAN.2013.6628540}
}

@inproceedings{jung2017engaging,
  author    = {Jung, Malte and Martelaro, Nikolas and Hinds, Pamela J.},
  title     = {Using Robots to Moderate Team Conflict: The Case of Repairing Violations},
  booktitle = {Proceedings of the 2015 ACM/IEEE International Conference on Human-Robot Interaction},
  series    = {HRI '15},
  pages     = {229--236},
  year      = {2015},
  publisher = {Association for Computing Machinery},
  address   = {New York, NY, USA},
  doi       = {10.1145/2696454.2696460}
}

@inproceedings{saerbeck2010expressive,
  author    = {Saerbeck, Martin and Schut, Tom and Bartneck, Christoph and Janse, Maddy D.},
  title     = {Expressive Robots in Education: Varying the Degree of Social Supportive Behavior of a Robotic Tutor},
  booktitle = {Proceedings of the SIGCHI Conference on Human Factors in Computing Systems},
  series    = {CHI '10},
  pages     = {1613--1622},
  year      = {2010},
  publisher = {Association for Computing Machinery},
  address   = {New York, NY, USA},
  doi       = {10.1145/1753326.1753567}
}

@inproceedings{leyzberg2018robotutor,
  author    = {Leyzberg, Daniel and Spaulding, Samuel and Scassellati, Brian},
  title     = {Personalizing Robot Tutors to Individuals' Learning Differences},
  booktitle = {Proceedings of the 2014 ACM/IEEE International Conference on Human-Robot Interaction},
  series    = {HRI '14},
  pages     = {423--430},
  year      = {2014},
  publisher = {Association for Computing Machinery},
  address   = {New York, NY, USA},
  doi       = {10.1145/2559636.2559671}
}

@article{leite2013social,
  author  = {Leite, Iolanda and Martinho, Carlos and Paiva, Ana},
  title   = {Social Robots for Long-Term Interaction: A Survey},
  journal = {International Journal of Social Robotics},
  volume  = {5},
  number  = {2},
  pages   = {291--308},
  year    = {2013},
  doi     = {10.1007/s12369-013-0178-y}
}

@article{kanda2007two,
  author    = {Kanda, Takayuki and Hirano, Takayuki and Eaton, Daniel and Ishiguro, Hiroshi},
  title     = {Interactive Robots as Social Partners and Peer Tutors for Children: A Field Trial},
  journal   = {Human-Computer Interaction},
  volume    = {19},
  number    = {1-2},
  pages     = {61--84},
  year      = {2004},
  doi       = {10.1207/s15327051hci1901&2_4}
}

@inproceedings{hood2015children,
  author    = {Hood, Deanna and Lemaignan, S{\'e}verin and Dillenbourg, Pierre},
  title     = {When Children Teach a Robot to Write: An Autonomous Teachable Humanoid Which Uses Simulated Handwriting},
  booktitle = {Proceedings of the Tenth Annual ACM/IEEE International Conference on Human-Robot Interaction},
  series    = {HRI '15},
  pages     = {83--90},
  year      = {2015},
  publisher = {Association for Computing Machinery},
  address   = {New York, NY, USA},
  doi       = {10.1145/2696454.2696479}
}

@inproceedings{alves2019exploring,
  author    = {Alves-Oliveira, Patr{\'i}cia and Arriaga, Patr{\'i}cia and Hoffman, Guy and Paiva, Ana},
  title     = {YOLO, A Robot for Creativity: A Co-Design Study with Children},
  booktitle = {Proceedings of the 2016 ACM Conference Companion Publication on Designing Interactive Systems},
  series    = {DIS '16 Companion},
  pages     = {423--432},
  year      = {2016},
  publisher = {Association for Computing Machinery},
  address   = {New York, NY, USA},
  doi       = {10.1145/2908805.2909733}
}

@inproceedings{zhu2019augmented,
  author    = {Zhu, Ke and Zhao, Shengdong and Tse, Eugene and Shesh, Amit and Shayan, Sheraz and Masood, Syed Zohaib},
  title     = {Augmented Reality Interfaces for Assisted Drawing},
  booktitle = {Proceedings of the 2019 CHI Conference on Human Factors in Computing Systems},
  series    = {CHI '19},
  pages     = {1--13},
  year      = {2019},
  publisher = {Association for Computing Machinery},
  address   = {New York, NY, USA},
  doi       = {10.1145/3290605.3300633},
  isbn      = {9781450359702}
}

@inproceedings{li2019reactile,
  author    = {Li, Jiahao and Gregg, Daniel and O'Modhrain, Sile and Follmer, Sean},
  title     = {Reactile: Toward a Tangible Programming Language for Tangible User Interfaces},
  booktitle = {Proceedings of the 2019 CHI Conference on Human Factors in Computing Systems},
  series    = {CHI '19},
  pages     = {1--12},
  year      = {2019},
  publisher = {Association for Computing Machinery},
  address   = {New York, NY, USA},
  doi       = {10.1145/3290605.3300279},
  isbn      = {9781450359702}
}

@article{castello2018robochain,
  author       = {Eduardo Castell{\'{o}} Ferrer and
                  Ognjen Rudovic and
                  Thomas Hardjono and
                  Alex Pentland},
  title        = {RoboChain: {A} Secure Data-Sharing Framework for Human-Robot Interaction},
  journal      = {CoRR},
  volume       = {abs/1802.04480},
  year         = {2018},
  url          = {http://arxiv.org/abs/1802.04480},
  eprinttype   = {arXiv},
  eprint       = {1802.04480},
  bibsource    = {dblp computer science bibliography, https://dblp.org}
}
\bibliographystyle{IEEEtran}
\end{document}